\documentclass[utf8]{frontiersSCNS}
\usepackage{url,hyperref,lineno,microtype,subcaption}
\usepackage[onehalfspacing]{setspace}
\usepackage{comment}
\def\keyFont{\fontsize{8}{11}\helveticabold }

\def\firstAuthorLast{Clabaugh {et~al.}}
\def\Authors{Caitlyn Clabaugh\,$^{1,*}$, Kartik Mahajan\,$^{1,*}$, Shomik Jain\,$^{1}$, Roxanna Pakkar\,$^{1}$, David Becerra\,$^{1}$, Zhonghao Shi\,$^{1}$, Eric Deng\,$^{1}$,  Rhianna Lee\,$^{1}$,  Gisele Ragusa\,$^{2}$, and Maja Matari\'{c}\,$^{1}$}
\def\Address{$^{1}$Interaction Lab, University of Southern California, Viterbi School of Engineering, Department of Computer Science, Los Angeles, California, USA \\
$^{2}$STEM Education Research Group, University of Southern California, Viterbi School of Engineering, Division of Engineering Education, Los Angeles, California, USA}
\def\corrAuthor{Kartik Mahajan}
\def\corrEmail{kmahajan@usc.edu}

\begin{document}
\onecolumn
\firstpage{1}

\title[Long-Term Personalization of a Home Robot for Children with ASD]{Long-Term Personalization of an In-Home Socially Assistive Robot for Children with Autism Spectrum Disorders}

\author[\firstAuthorLast ]{\Authors} 

\address{} 
\correspondance{} 
\extraAuth{ *The two leading authors contributed equally to this paper.}

\maketitle

\begin{abstract}

Socially assistive robots (SAR) have shown great potential to augment the social and educational development of children with autism spectrum disorders (ASD). As SAR continues to substantiate itself as an effective enhancement to human intervention, researchers have sought to study its longitudinal impacts in real-world environments, including the home. Computational personalization stands out as a central computational challenge as it is necessary to enable SAR systems to adapt to each child's unique and changing needs. 
Toward that end, we formalized personalization as a hierarchical human robot learning framework (hHRL) consisting of five controllers (disclosure, promise, instruction, feedback, and inquiry) mediated by a meta-controller that utilized reinforcement learning to personalize instruction challenge levels and robot feedback based on each user's unique learning patterns. We instantiated and evaluated the approach in a study with 17 children with ASD, aged 3 to 7 years old, over month-long interventions in their homes. Our findings demonstrate that the fully autonomous SAR system was able to personalize its instruction and feedback over time to each child's proficiency. As a result, every child participant showed improvements in targeted skills and long-term retention of intervention content. Moreover, all child users were engaged for a majority of the intervention, and their families reported the SAR system to be useful and adaptable. In summary, our results show that autonomous, personalized SAR interventions are both feasible and effective in providing long-term in-home developmental support for children with diverse learning needs.

\tiny
 \keyFont{ \section{Keywords:} long-term human-robot interaction, personalization, socially assistive robotics, reinforcement learning, home robot, autism spectrum disorders, early childhood}
\end{abstract}

\section{Introduction}

Human development follows nonlinear trajectories unique to each individual \citep{vygotsky1978interaction}. Therefore, socially assistive interventions need to be tailored towards the specific needs and preferences of each participant over time. In a long-term setting, this means interventions must continuously and rapidly adapt toward the user's unique personality. Given the complexity, unpredictability, and uniqueness of each user's progress, intervention strategies must be adapted {\it in situ} via untrained human feedback. Creating autonomous long-term personalized adaptation poses many computational and engineering challenges.

Benefits of personalization are well established across the domains of education \citep{bloom19842,anderson2001taxonomy} and healthcare \citep{beevers2012therapygenetics,cesuroglu2012public,swan2012health}. While personalized services are paramount, they are neither universally nor equitably affordable. This provides motivation for human-machine interaction research that seeks to develop personalized assistance via socially assistive agents, whether disembodied, virtually embodied \citep{devault2014simsensei}, or physically embodied \citep{Mataric2017Sociallyassistive}.

Socially assistive robotics (SAR) combines robotics and computational methods to broaden access to personalized, socially situated, and physically co-present interventions \citep{feil2011socially}. A large body of work has supported the importance of physical embodiment \citep{deng2019Embodiment}, including its role in increasing compliance \citep{bainbridge2008effect}, social engagement \citep{wainer2006role,lee2006physically}, and cognitive learning gains \citep{leyzberg2012physical}. Correspondingly, there has been a significant body of work using various types of robots for children with autism spectrum disorders (ASD) in short-term studies \citep{diehl2012clinical, scassellati2012robots, begum2016robots}, and one long-term study \citep{Scassellati2018improving}. 

The majority of past work with SAR for ASD has been related to social skills.  However, it is well established that learning in general is impacted by social factors; this is particularly important for young learners, because their learning is most often socially mediated \citep{durlak2011}.  Social difficulties often interfere with children’s learning; therefore embedding social contexts in learning environments presents a developmentally appropriate practice that is preferable over isolating social behaviors from cognitive activities \citep{zins2004}.  Consequently, this work addresses the social and cognitive learning domains in tandem, in an intervention that is specifically designed for such learning by children with ASD \citep{guadalupe2016,white2007social}. 

Personalizing the learning process is especially important in ASD. Given sufficient domain knowledge, personalization of SAR can be achieved through human-in-the-loop or Wizard of Oz (WoZ) frameworks, wherein intervention strategies are mapped to individuals {\it a priori} or {\it in situ} via human input \citep{riek2012wizard}. However, in practice, considering diverse individual needs and the noise of real-world environments, and the scale of need in ASD, non-autonomous personalization of SAR is infeasible. Reinforcement learning (RL) methods have been successfully applied to adapting to a user's learning habits over time, particularly in early child development studies \citep{ros2011child}. Moreover, recent long-term SAR studies have demonstrated success in maintaining persistent co-present support for educators, students, and caregivers \citep{bongaarts2004population}. There is therefore an opportunity to develop RL-based personalized long-term learning SAR systems, especially when teaching abstract concepts such as mathematics  \citep{clabaugh2015designing}.

In this work, we propose a {\it personalized} SAR intervention framework that can provide accessible and effective long-term, in-home support for children with ASD. To accommodate the variable nature of ASD, our framework personalizes to each user's individual needs. To that end, we introduce a hierarchical framework for Human Robot Learning (hHRL) that decomposes SAR interventions into computationally tractable state-action subspaces contained with a meta-controller. The meta-controller consists of disclosure, promise, instruction, feedback, and inquiry controllers that personalize instruction challenge levels and robot feedback based on each child's unique learning patterns. The framework is implemented and evaluated in a fully autonomous SAR system deployed in homes for session-based, single-subject interventions with 17 child participants diagnosed with ASD aged 3 to 7 years old. Using space-themed mathematics problems, the system combined tenets of educational robotics and computational personalization to maximize each child participant's cognitive gains.  Our findings show that the SAR system successfully personalized its instruction and feedback to each participant over time. Furthermore, most families reported the SAR system to be useful and adaptable, and correspondingly, all users were engaged for a majority of the in-home intervention. As a result, all participants showed improvements in math skills and long-term retention of intervention content. These outcomes demonstrate that computational personalization methods can be successfully incorporated in long-term personalized SAR deployments to support children with diverse learning needs. 

This paper is organized as follows. \textit{Background} overviews SAR in the relevant contexts of learning, ASD, and personalization. \textit{Formalizing Personalization in SAR} describes the hierarchical human robot learning framework, with a focus on personalization of the challenge level and robot feedback. \textit{Personalized SAR Intervention Design} details the study design, data collection, and outcome measures. The \textit{Results} section details the adaptation performance of the SAR system, its influence on user engagement, participating families' perspectives, and cognitive learning gains over the long-term interaction. \textit{Discussion} and \textit{Conclusion} summarize key insights and recommendations for future work. 
\section{Background}

Socially Assistive Robotics (SAR) lies at the intersection of socially interactive robotics and assistive robotics, and focuses on developing intelligent, socially interactive robots that provide assistance through social interaction, with measurable outcomes \citep{feil2005defining, Mataric2016}. We review the relevant background in the main contribution areas of this work: SAR for \textit{learning} (Section~\ref{sar:learn}) and SAR for \textit{personalization} (Section~\ref{sar:personalization}), both with a particular emphasis on the ASD context, given particular challenges and opportunities for SAR.

\subsection{SAR for Learning}\label{sar:learn}
A large body of evidence across multiple disciplines supports personalized instruction as a means of positively impacting development and motivation of {\it individual} learners. Examples include personalized tutoring systems in human-computer interaction research \citep{wenger2014artificial}, personalized robot tutors in HRI and SAR research \citep{leyzberg2014personalizing}, and optimal challenge points \citep{guadagnoli2004challenge} and the Zone of Proximal Development methodologies in education research \citep{chaiklin2003zone}. 

A significant body of SAR research has focused on user learning, with a specific focus on developing personalized robot tutors for young children \citep{Clabaugh2019Escaping}. Many SAR and HRI studies have found a robot's embodiment to augment learning in a variety of settings \citep{gallese1998mirror, lee2006physically, gazzola2007anthropomorphic, wainer2007embodiment, bainbridge2008effect,leyzberg2012physical,fridin2014embodied}. Additionally, several studies on intelligent tutoring systems (ITS) have involved computational models of student learning patterns; however, in contrast to SAR, these works have predominately focused on university students in highly controlled environments \citep{anderson1985cognitive, murray1999authoring}. From that body of past work, key principles about SAR for learning have been grounded in theories of embodied cognition, situated learning, and user engagement. 

\textit{Embodied cognition} research has shown that knowledge is directly tied to perceptual, somatosensory, and motoric experience, and that a robot's physical embodiment can help contextualize a user's ideas \citep{niedenthal2007embodying, deng2019Embodiment}. For example, SAR has helped participants develop motor \citep{goldin2010action}, behavioral \citep{fong2003survey} and cognitive skills \citep{toh2016review}. SAR has also shown success in helping users learn abstract concepts; for example, \cite{clabaugh2015designing} implemented a SAR system that used deictic gestures to help preschoolers learn number concepts. 

\textit{Situated learning} refers to the importance of the social and physical environment on the learning process and outcomes \citep{mclellan1996situated}. Cognitive gains are dependent on context and are enhanced by social interaction \citep{anderson1996situated}. Therefore, SAR intervention efficacy must be analyzed in real-world learning settings, involving user learning in various spatial and social contexts \citep{sabanovic2006robots}. Environmental conditions impact the quality of SAR interactions and the resulting assistive outcomes. However, real-world scenarios are inherently noisier and less predictable, requiring more complex experimental designs and robust robot platforms \citep{ros2011child}. 

\textit{User engagement} is an important measure of SAR's effectiveness and is inherently tied to learning. In the context of HRI, engagement is widely accepted as a combination of behavioral, affective, and cognitive constructs. Specifically, engagement involves on-task behavior, interest in the robot and task at hand, and a willingness to remain focused \citep{scassellati2012robots}. \cite{Rudovic2018Personalized} successfully modeled users’ engagement with a personalized deep learning framework, however the model was developed {\it post hoc}, not in real time. As discussed in \cite{Kidd2008Designing}, maintaining user engagement in real time is a major challenge for real-world, long-term studies, as are overcoming technological difficulties and accounting for external human actors. 

All of the challenges of SAR for learning are significantly amplified in the ASD context, but ASD is also the context where the success of SAR in supporting learning is especially promising. ASD is a complex developmental disorder that is often marked by delays in language skills and social skills, including turn-taking, perspective-taking, and joint attention \citep{white2007social}. Personalized therapeutic and learning interventions are critical for individuals with ASD, but the substantial time and financial resources required for such services make them inaccessible to many \citep{ospina2008behavioural, lavelle2014economic}, creating an opportunity for SAR support. 

There is a large and growing body of research on using SAR for a variety of ASD interventions, as reviewed in \cite{diehl2012clinical, scassellati2012robots, begum2016robots}. SAR has been shown to help children with ASD develop behavioral and cognitive skills, specifically increased attention \citep{Duquette2008Exploring}, turn-taking \citep{baxter2013emergence}, social interaction \citep{robins2005robotic}, and many other skills.  SAR's ability to perceive, respond, and adapt to user behavior is especially critical in the ASD context \citep{Clabaugh2019Escaping}, as users with ASD vary greatly in symptoms and severities, underscoring the need for personalization, as our work also demonstrates.


\subsection{Personalization in SAR}\label{sar:personalization}
SAR systems have shown great potential for providing long-term situated support for meeting individual learning needs.  Autonomous or computational personalization in SAR often seeks to maximize the participants' focus and performance, using rule-, model-, or goal-based approaches to personalization.

\textit{Rule-based} approaches to personalization have been successful in both short-term and long-term SAR interventions. For example, \cite{Ramachandran2018} designed single session interventions where the robot encouraged participants to think out loud. Scenarios were presented based on whether a participant successfully answered a question, and this simple rule-based method resulted in learning gains across all users. Additional studies have expanded rule-based approaches for sequential interactions using hierarchical decision trees \citep{Kidd2008, Reardon2015}. Furthermore, in a study setup similar to ours, \cite{Scassellati2018improving} developed a personalized SAR system for month-long interventions with children with ASD. The system adapted the challenge level of activities using past performance and fixed thresholds. As a result, participants showed increased engagement to the robot and improved attention skills with adults when not in the presence of the robot. In contrast, this work personalizes feedback and challenge level using a goal-based approach, discussed below.

\textit{Model-based} approaches use models to evaluate the user's success and make optimal decisions. Bayesian Knowledge Tracing (BKT), a domain-specific form of Hidden Markov Models (HMMs), is a common \textit{model-based} approach to personalization in SAR where the hidden state is based on the user's performance and loosely represents their knowledge \citep{desmarais2012review, van2013properties}. For example, BKT can assess how well a participant understands a concept such as basic addition by examining the sequence of the user's correct and incorrect responses. To represent the variability present in most learning interactions, BKT uses two domain-specific parameters: the probability that a participant will slip and the probability that they will guess. These parameters are dependent on the interaction context; students with ASD may have difficulty concentrating for extended periods and thus may slip more frequently than typically-developing users \citep{schiller1996educating}.  BKT has been successfully applied in SAR; \cite{Gordon2015} and \cite{Schodde2017} used it to adapt to user age and experience, leading to increased learning gains. \cite{Leyzberg2014} applied BKT to training a SAR system to help users solve challenging puzzles more quickly. While outside of ASD, these studies demonstrate the value of BKT in adapting SAR to varying learner needs. 

\textit{Goal-based} methods help the SAR system to select actions that maximize the user’s progress toward an assistive outcome. Reinforcement Learning (RL) is a popular goal-based approach, where each user action produces some reward representing progress toward the goal. Throughout the interaction, RL develops a unique, personalized strategy for each participant based on reward-favoring paths. Within HRI, RL has been used to maximize the user's affective state, leading to more effective interactions \citep{Conn2008, Chan2011, Castellano2012, Gordon2016}. Prior studies have shown RL to require deep datasets given the noise of real-world environments. In a single-session context, \cite{Gordon2016} showed RL to successfully adapt in an average of three out of seven sessions. \cite{Castellano2012} also utilized RL to increase engagement; the model was trained on a 15 minute interaction and was no better at adapting than a randomized empathetic policy.  \cite{Conn2008} showed a RL which was able to adapt quickly, but simplified the robot state to three distinct behaviors. To enable a broader range of behaviors, \cite{Chan2011} implemented a hierarchical RL model to personalize feedback within a memory-based SAR interaction.  As demonstrated by past work, long-term studies provide the datasets needed for effective RL-based personalization.

Related work has addressed improving social skills of children with ASD. To manage noisy environments and the unpredictable nature of ASD, two studies are particularly relevant as they used RL to parameterize action spaces and speed up robot learning. \cite{Veletzas2018} used RL to personalize the robot's actions to maximize a child's engagement; the robot guided children through a Tower of Hanoi puzzle and used RL to effectively identify nonverbal cues and teach at the learning rate of the participant. That work parameterized the robot's action space to enable efficient decision making and learn single moves in the absence of traditional hierarchical models. \cite{Khamassi2018} utilized a parameterized action space to select the appropriate robot reaction that maximizes a child's engagement. That work also used RL to maximize the participant's engagement when interacting with a robot. By using the participant's gaze and past variations in engagement, their Q-Learning algorithm became more robust over time. The work used a parameterized environment to simultaneously explore a discrete action space (e.g., moving an object) and a continuous stream of movement features (e.g., expressivity, strength, velocity). These two studies provide insight into maximizing engagement in the absence of hierarchical models, especially when encouraging social interaction (e.g., talking, moving). 

The work described in this paper is complementary but different from past work in that it analyzes a long-term SAR intervention for abstract concept learning, specifically helping children with ASD learn mathematics skills. As the next section details, a goal-based RL approach was developed to personalize the instruction and feedback provided to each child by the SAR system.
\section{Formalizing Personalization in SAR} \label{mm:per}
To address the challenge of long-term personalization in SAR in a principled way, we present a solution to the problem as a controller-based environment which we define as hierarchical human-robot learning (hHRL).    

\subsection{Human-Robot Learning} \label{mm:per:hrl}
Past work has explored methods for computational personalization, with the objective of finding an optimal sequence of actions that steers the user toward a desired goal. While this problem has been studied in the contexts of user modeling in HCI \citep{fischer2001user}, machine teaching \citep{chen2018understanding}, as well as active \citep{cohn1996active} and interactive machine learning (ML) \citep{amershi2014power,dudley2018review}, computational personalization is yet to be formalized in the context of SAR. 

We define and formalize {\it Human-Robot Learning (HRL)} as the interactive and co-adaptive process of personalizing SAR. At the highest level, the quality of a SAR intervention can be assessed relative to some goal \(G\). Since SAR contexts often involve long-term goals, success is better assessed via intermediate measures of progress toward \(G\). Hence, it is important to design and represent SAR intervention interactions in a manner that maximizes observability. In this work, HRL is framed from the perspective of the robot, so that optimization is limited to the robot's actions and not those of the human, in contrast to human-robot collaboration and multi-agent learning \citep{nikolaidis2016formalizing,littman1994markov}. 

\subsection{Hierarchical HRL} \label{mm:per:hierarchy}
\begin{figure} 
    \centering
    \includegraphics[width=0.8\linewidth]{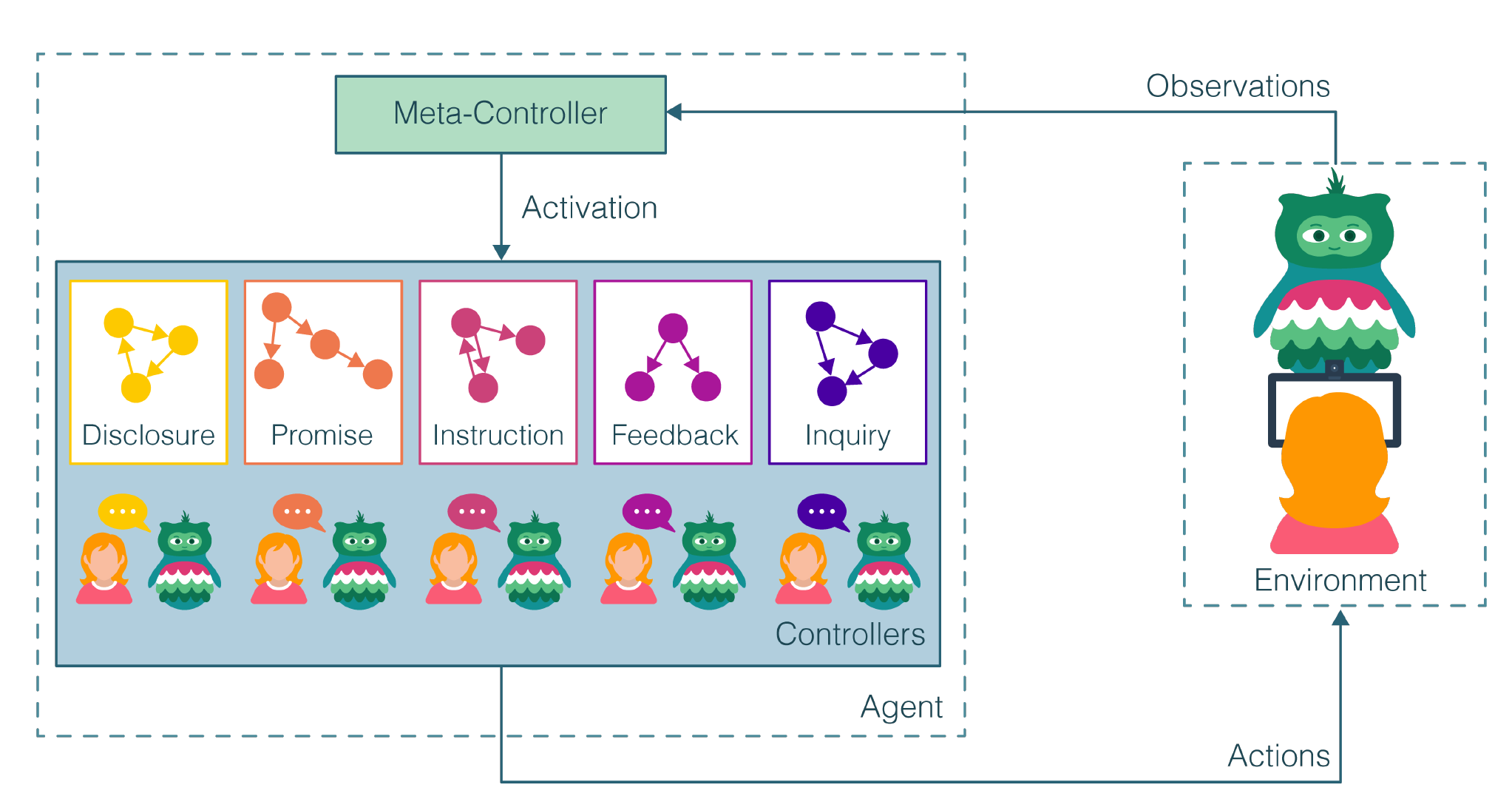}
    \caption{The hierarchical framework for human-robot learning (hHRL) comprises of a two-level hierarchy: (1) The \textit{meta-controller} takes high-level information about the current state of the intervention and activates a lower-level controller. (2) The lower-level controllers await activation to select the robot's action based on a simplified state representation, reward, and action category: instructions, promises, feedback, disclosures, and inquiries.}
    \label{abcontrollers}
\end{figure}

We introduce {\it a hierarchical framework for HRL (hHRL)} as one that decomposes SAR interventions into computationally tractable state-action subspaces. Building on the work of \cite{kulkarni2016hierarchical}, the hHRL framework is structured as a two-level hierarchy, shown in Figure \ref{abcontrollers}. At the top level, a meta-controller considers high-level information about the intervention state and activates some lower-level controller. SAR-specific controllers wait for activation to select the robot's action based on a simplified state representation. The hHRL framework assumes that SAR interventions can be characterized by five abstract action categories: 1) instructions \(I\), 2) promises \(P\), 3) feedback \(F\), 4) disclosures \(D\), and 5) inquiries \(Q\). Each category is modeled by a separate controller activated by the overarching meta-controller. 

Each controller is responsible for a theoretical subset of SAR actions, henceforth referred to as \textit{SAR acts}. In this work, SAR acts are formalized based on the directive, commissive, and representative illocutionary speech acts, or simply illocutions, originally defined in linguistic semantics \citep{austin1975things,searle1969speech}. \cite{searle1976classification} defined {\it illocutions} in terms of speaker, hearer, sincerity condition, psychological state, propositional content, and direction of fit. In the SAR context, the speaker is the robot, the hearer is the human user, and the direction of fit is either action-to-state, where the objective is to make the robot's action match the state of the intervention, or state-to-action, where the objective is to make the state of the intervention match what is expressed through the robot's action. Illocutions uniquely manifest themselves through other modes of communication, such as gestures \citep{mehrabian2017nonverbal}, pictures \citep{danesi2016semiotics}, music \citep{kohn2004treatment}, and other multimodal signals \citep{forceville2009multimodal, horn1998visual}. These alternative signals are particularly relevant to SAR because robots have inherently expressive embodiments \citep{fong2003survey}. Additionally, SAR interventions target special populations \citep{feil2005defining}, such as linguistic minorities (e.g., American Sign Language \citep{stokoe1976dictionary}) or personals with disabilities that involve speech and language difficulties or delays (e.g., Dysarthria \citep{darley1969differential} or autism spectrum disorder \citep{kasari2012making}). Therefore, SAR acts are defined to be illocutions irrespective of communicative modality. 

\subsection{Abstract Controllers} \label{mm:per:abstract controllers}
Within the hierarchical model, {\it instructions} are defined as attempts by the robot to get the user to do something that might generate progress toward the intervention goal. Within the instruction controller, there may be some predefined or learned ordering among instructions, such as the level of challenge or specificity.

{\it Feedback} is defined as beliefs expressed to the user by the robot about their past and current interactions. The {\it direction of fit} is action-to-state, the {\it sincerity condition} is belief \(B\), and the {\it propositional content} is that some past or current state \(s\) had or has some property \(p\). In this way, feedback \(F\) is defined as a specific form of representatives. {\it Representatives} were defined by \cite{searle1976classification} to commit the speaker to the truth of the expressed proposition. The {\it propositional content} is information about the state relative to some instruction or goal. The feedback controller is responsible for selecting the information or assistance given to the user by the robot. Feedback can be modeled in a variety of ways, the impacts of which have been studied in psychology and human-machine interaction. Specifically, feedback can be adapted to match individual proficiency or independence, as in scaffolded \citep{finn2010scaffolding} or graded cueing models \citep{feil2012simon,greczek2013computational}. It can also be modeled to increase self-efficacy, as in the growth mindset \citep{o2014brain,park2017growing} and constructive feedback models \citep{ovando1994constructive}. Additionally, feedback timing has also been studied \citep{kulik1988timing}, such as feedback in response to help-seeking \citep{roll2011improving} and disengagement \citep{leite2015comparing}. 

Extrinsic motivation is a well-studied driver of behavior, explored in educational \citep{vallerand1992academic}, professional \citep{amabile1993motivational}, and personal settings \citep{sansone2000intrinsic}, as well as a common measure in evaluating the effectiveness of human-robot interaction \citep{fasola2012using,breazeal1998early,dautenhahn2007socially}. In the hHRL framework, {\it promises} are defined as commitments made by the robot for performing future actions that aim to motivate the user through the promise controller. Promises also relay information critical to collaboration and transparency, expressed via verbal or nonverbal signals such as gross motion \citep{dragan2013legibility}. Although they are not directly tied to quantitative measures, promises help to make the robot more personable and consistent over a long-term study period.

{\it Disclosures} are defined as beliefs expressed to the user by the robot about its past or current self. The disclosure controller selects internal information for the robot to share with the user as a means of fostering human-robot reciprocity and solidarity. Robot transparency has shown to increase trust \citep{hancock2011meta,yagoda2012you}, improve collaboration \citep{kim2006should,breazeal2005effects}, and build empathetic relationships \citep{leite2013influence}. Past work has also shown that nonverbal signals can be particularly effective in disclosing internal states such as emotion  \citep{bruce2002role}. 

{\it Inquiries} are defined as attempts by the robot to get the user to express some truth. The {\it inquiry controller} selects what information the robot should attempt to elicit. Inquires may be posed for a variety of interaction benefits, such as improving engagement, relationship, and trust \citep{hancock2011meta}. Inquires may also be used to gather feedback about the robot or information about the user, as in interactive machine learning \citep{amershi2014power}. 

 \begin{figure}[h!]
    \centering
    \includegraphics[width=0.4\textwidth]{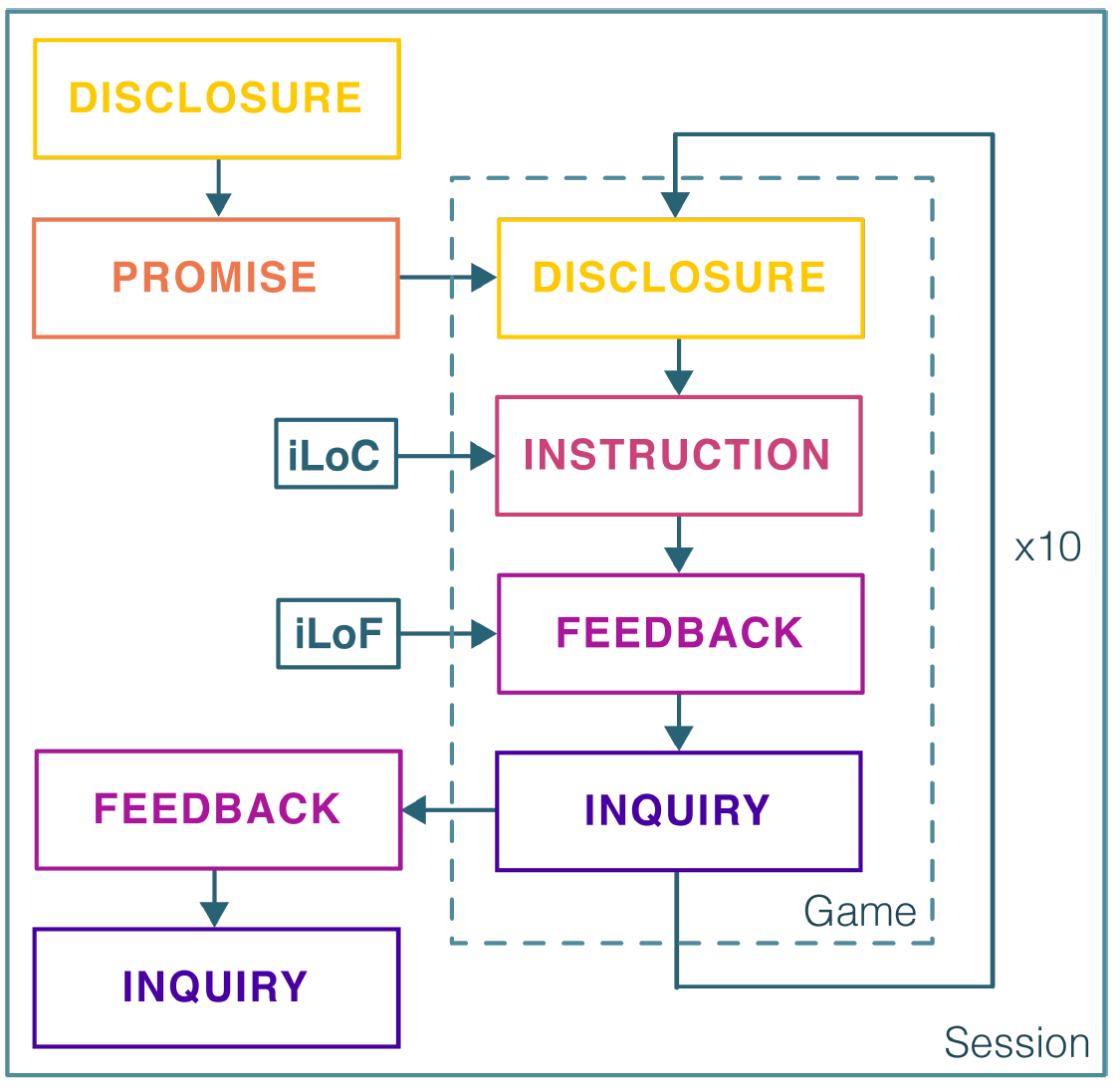}
     \caption{The real-world long-term SAR intervention for early childhood math learning used a meta-controller that sequentially executed each controller. At the beginning of a session, the robot {\it disclosed} that it needed the child's help to reach a specific planet. It then {\it promised} that they would reach the planet if they completed all the games they needed to do that day. The child and robot played 10 games whose challenge (pLoC) and feedback (pLoF) levels were set by the computational personalization methods. At the end of each session, the robot congratulated the child on completing the games and reaching the promised planet. It then asked the child some open-ended questions about their day.}
     \label{fig-instantiation}
 \end{figure}

\subsection{Computational Personalization} \label{mm:per:instantiation}

To personalize the SAR system, the proposed hHRL controller was instantiated as a group of domain controllers, based on the abstract controllers defined in Section \ref{mm:per:abstract controllers}. Figure \ref{fig-instantiation} represents how the abstract controllers were contained within a domain-specific \textit{meta-controller}. The meta-controller activated one controller at a time.  We used insights and data from our prior work, reported in \cite{clabaugh2015designing}, to inform the design of the controllers for SAR personalization.  Our prior study collected data from 31 typically developing preschool children who interacted with a SAR tutor in a single session at their child development center preschool. The data were used to develop a model that predicted a child's performance on the game. We used this performance model to bootstrap the prediction of the children’s performance in our study.
Specifically, the instruction controller in the personalization framework optimized the level of challenge (pLoC) and the level of feedback (pLoF) to match each child’s performance, as described next. 

\subsubsection{Personalization of the Level of Challenge}\label{mm:per:challenge}

Personalization was partially accomplished within the instruction controller. Learning games $g$ were randomly sampled without replacement from all games $G$ and parameterized by some personalized {\it level of challenge (LoC)} $c \in [1,5]$. The instruction controller was designed to optimize LoC to match individual proficiency. This optimization problem was based on the concepts of \textit{optimal challenge} from the \textit{Challenge Point Framework} by \cite{guadagnoli2004challenge} and from the research on the \textit{Zone of Proximal Development} (ZPD) by \cite{chaiklin2003zone}; both define the goal as challenging individuals enough that they are presented with new information, but not so much that there is too much new information to interpret.

Since the goal is long-term adaptation, {\it personalized LoC (pLoC)} was framed as a RL problem, trained using Q-learning \citep{watkins1992q}. Within the instruction controller, a reward function was used to quantify the intervention state and supply Q-learning. The intervention, at time $t$, was defined by:
\begin{enumerate}
    \item the current game $g_t$,
    \item the current LoC $c_t$, and
    \item the current number of mistakes $m_t$
\end{enumerate}
More formally, the state space was defined as $G$ and the action space was defined as $C$, for a total of $G \times C = 10 \times 5 = 50$ (state, action) couples. As previously explained, the next game $g$ was randomly sampled without replacement from all games $G$. Therefore, the RL seeks to find a policy with the optimal LoC $c \in [1,5]$ per game for the individual child.

Given the formulation above, the RL would select and evaluate different LoCs for each child. If some LoC in some game was too difficult or too easy for a child, the RL would learn to select a different LoC for that game, over time. This was accomplished through a reward function designed to maximize LoC without pushing the learner to make too many mistakes. Formally, at time $t$, let $m_t$ be the number of mistakes a learner has made and $M$ be the predefined threshold of maximum mistakes (we used a threshold of five, based on our empirical findings from prior research \citep{clabaugh2015designing}). The reward function $R(t)$ returns a value equivalent to the LoC $c_t$, unless the $m_t > M$; then, $R(t)$ returns the inverse of LoC.

\begin{equation}\label{eq:loc}
R(t) = c_t \cdot MC(t),
\end{equation}
where
\begin{equation}\label{eq:loc-mistakes}
  MC(t)=\begin{cases}
    1, & \text{if $m_t \leq M$}\\
    -1, & \text{otherwise}.
  \end{cases}
\end{equation}\label{eq:thresh}

\subsubsection{Personalization of the Level of Feedback}\label{mm:per:feedback}

\cite{feil2012simon} and \cite{greczek2014graded} applied the concept of {\it graded cueing} to adapt feedback in the context of SAR interventions for children with ASD. A similar approach was taken in this work to instantiate the feedback controller, as mentioned in Section \ref{mm:per:instantiation}.  Analogous to the instruction controller, the feedback controller was modeled as a MDP, wherein the decision was to select one of five levels of feedback (LoF) $f \in [1,5]$ to match individual need. The feedback actions were specific to early mathematics learning. 

{\it Personalized LoF (pLoF)} was framed as a RL problem, trained using Q-learning \citep{watkins1992q} over many repeated interactions. Within the feedback controller, at time $t$, the intervention was represented by four parameters:
\begin{enumerate}
    \item the current game $g_t$,
    \item the current LoF $f_t$,
    \item the current number of mistakes $m_t$, and
    \item the current number of help requests $h_t$
\end{enumerate}
Similar to LoC, the state space for the feedback controller consisted of the $G = 10$ game states. The action space consisted of the four LoFs $f \in [1,4]$. The final LoF $f=5$ was not included as part of the personalization problem. The final feedback level was selected if and only if the child made more than the five allotted mistakes, and the meta-controller would move on to the next interaction. Therefore, the feedback controller included a total of $G \times F = 10 \times 4 = 40$ (state, action) couples.

The reward function was designed to minimize LoF without pushing the learner to make too many $M$ mistakes (where $M$ was the predefined threshold of maximum mistakes) or penalizing them too heavily for making help requests.

\begin{equation}\label{eq:lof}
R(t) = (-1 \cdot \frac{f_t}{m_t + h_t + 1}) + MC(t),
\end{equation}
where
\begin{equation}
  MC(t)=\begin{cases}
    5, & \text{if $m_t \leq M$}\\
    0, & \text{otherwise}.
  \end{cases}
\end{equation}\label{eq:thresh}
\section{Personalized SAR Intervention Design}
The SAR personalization framework was instantiated in a SAR systems designed for and evaluated in a month-long, in-home SAR intervention in the homes of children with ASD, and approved under USC IRB UP-16-00755.  The details of the SAR system, study design, data collection, and outcomes measures are described next.

\subsection{System Design} \label{mm:design}
The physical robot was designed to be a near-peer learning assistant, intended to act as the child's companion rather than tutor. Toward that end, it was given a neutral, nonthreatening character that presented educational games on a tablet and provided personalized feedback.

\subsubsection{Physical Design} \label{mm:design:pd}

\begin{figure}[t!]
    \centering
    \includegraphics[width=0.5\linewidth]{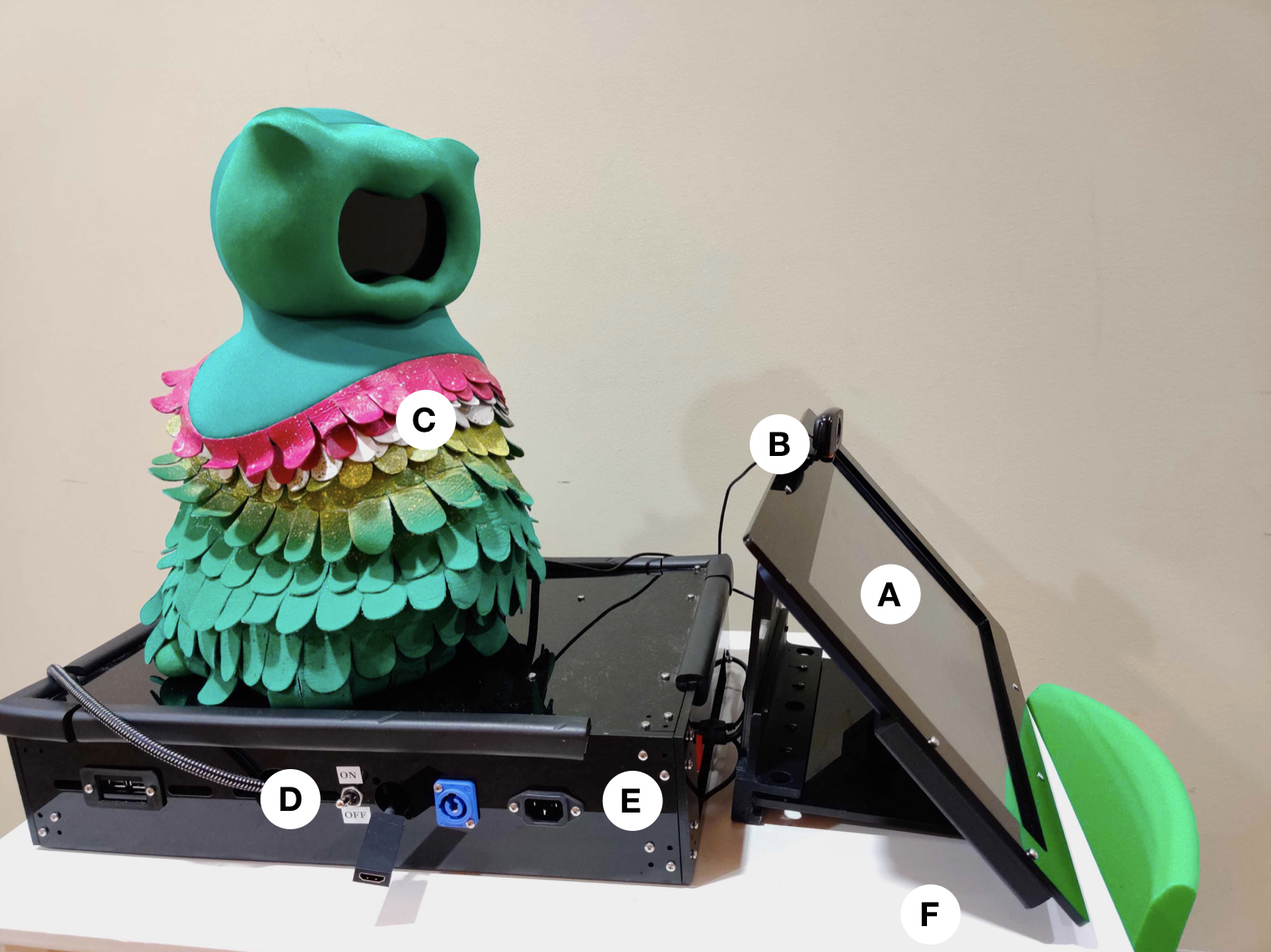}
    \caption{The physical in-home setup included the SPRITE robot with the Kiwi skin (C) mounted on top of the container encasing a computer, power supply, and speakers (E), with an easy-access power switch (D), a camera (B), and touchscreen monitor (A), all located on a standard child-sized table (F).}
    \label{system}
\end{figure}

To enable long-term in-home deployments, including ensuring the protection of the system's sensitive components, we designed a self-contained and portable system, shown in Figure \ref{system}, consisting of the robot, and a container that encompassed the robot's power supply, speakers, and tablet.  The container was approximately the same width as the robot to minimize the overall system footprint. 

The robot platform we designed was modified the Stewart Platform Robot for Interactive Tabletop Engagement (SPRITE) with the Kiwi skin \citep{short2017sprite}. SPRITE used CoR-Dial, also known as the Co-Robot Dialogue system, the software stack that controls the robot's physical movements and virtual face. The SPRITE consists of a 3D printed base, housing electronic components and threaded rods that support a laser-cut platform with six degrees of freedom. Within the exterior skin, a small display was used to animate the robot's face that included two eyes, eyebrows, and a mouth, all of which were controlled using Facial Action Coding System (FACS) coding in CoR-Dial.

The Kiwi skin and character were designed to appeal to the target user population. Children with ASD are often overwhelmed by sensory input, so Kiwi was designed to be non-threatening and simple in its affective displays. It was also gender-neutral in its appearance, allowing each child to assign the robot's gender if and as desired.  

\subsubsection{Game Design} \label{mm:design:games} 
The design of the SAR intervention was conceptualized by our multidisciplinary team of researchers, leveraging established game design principles, including iterative prototyping \citep{adams2013}. Through these processes, \cite{Clabaugh2018Attentiveness} designed an intervention that balanced the needs of the domain with the limitations of SAR technology. The initial game prototype was presented to a focus group of early childhood educators who served as subject matter experts and provided formative feedback on what was developmentally appropriate for children with ASD diagnoses. This informed the second generation of the game design, which was then piloted in a preschool classroom. Following these pilot studies, further adjustments were made to accommodate specific needs of children with ASD before the system was iteratively deployed and validated over multiple, long-term, in-home interventions for 17 children with ASD.

The game types within the system were tailored specifically for children with ASD, based on previous case studies, developmentally appropriate practices in working with children with ASD \citep{coppleandbredekamp2009}, and standards recognized by the National Association of the Education of Young Children. More specifically, they were developed in concert with developmentally appropriate practices for young children ages 3-8 and informed by contemporary learning theory \cite{omrod2017}. Each game employed a scaffolded approach to gradually increasing difficulty level as the child navigated through successful completion of a particular game level \cite{sweller2007}. Both the content and difficulty levels were also aligned both to best practices in child development standards of the National Association of Education of Young Children (NAEYC) and the National Common Core Mathematics Standards (for more advanced levels; CCMS \citep{coppleandbredekamp2009}).

The games were also aligned with the Wechsler Individual Achievement Test (WIAT), a developmental-level standard assessment \citep{wechesler2005}, used as a pre-post measure of the impact of the game, as described in Section \ref{mm:methods:measures:objective}. Numerical operation and math reasoning were selected as pre-academic content for the games because they are the early math skills needed by children in preschool and kindergarten. As a result, they are also areas that control for potential social biases found in many early childhood games.  

\begin{figure}[t!]
    \centering
    \includegraphics[width=0.7\linewidth]{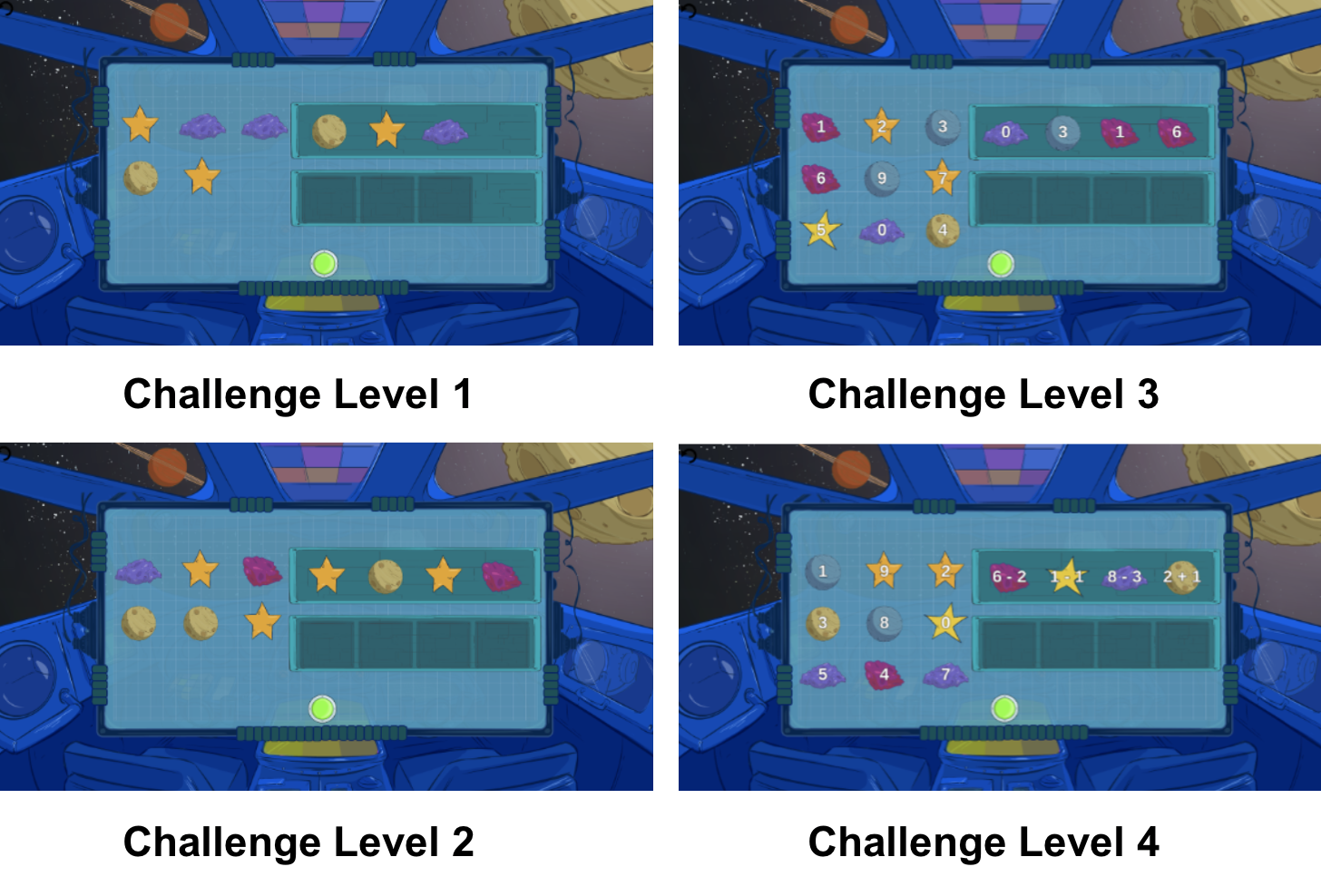}
    \caption{The child-robot interaction was designed around Kiwi as a robot space explorer. The following diagram displays varying challenge levels of the Pack Moon-Rocks game, with more challenging problems combining math reasoning and numerical operation concepts.}
    \label{games}
\end{figure}

Figure \ref{games} illustrates an example of the different challenge levels of one of the games; game challenge levels were personalized to each child participant as described in Section \ref{mm:per:challenge}.

\subsubsection{Child-Robot Interaction} \label{mm:design:ci}
The Kiwi character described itself as a space explorer and a peer to the child user that continuously needed help from the child in order to return to its home planet. Users were told they could help Kiwi by playing the provided tablet-based games. The games tested a variety of preschool and kindergarten math skills, including addition, counting, and pattern matching.  The SAR system offered ten different types of games based on five different levels of challenge (LoC). Child participants were encouraged to play at least one game during each interaction; the games involved the user performing the following on-screen tasks:
\begin{enumerate}
\item Pack Moon-Rocks: Drag 1-10 moon-rocks into a box. 
\item Select Galaxy: Select the galaxy with more or fewer stars.
\item Select Planet: Select the planet with a particular number.
\item Feed Space Pets: Evenly divide a set of stars between two “alien pets".
\item Pets on a Spaceship: Drag numbered “alien pets” into a spaceship in increasing or decreasing order.
\item Organize Moon-Rocks: Separate and organize moon-rocks based on sprite and number.
\item Organize Space Objects: Separate and organize various space-themed objects based on sprite and number.
\item Pattern Completion: Complete a pattern with the provided space objects.
\item Identify Alien Emotion: Determine the emotion of one or more “alien friends” based on their facial expressions.
\end{enumerate}

The graphics in the game used an age-appropriate comic book design style, with colorful aliens guiding the user through the games. Each game allowed up to five mistakes; every mistake was followed by a verbal hint delivered by Kiwi paired with child-like body movements that signaled  whether the user was struggling or excelling. The feedback actions were specific to the game context of early mathematics learning. For example, if the child was presented with the instruction ``Put five energy crystals into a box'' but used too few crystals, the feedback controller executed one of the following actions:
\begin{enumerate}
    \item ``We need to have a total of five energy crystals inside the box.''
    \item ``Try counting out loud as you drag each crystal one by one.''
    \item ``You have too few energy crystals. Try adding some to the box.''
    \item ``We currently have three energy crystals. So we need two more energy crystals. Can you drag two more crystals into the box?''
    \item ``Let's try something else.''
\end{enumerate}

\subsection{Study Participants} \label{mm:methods:participants} 
Seventeen children with ASD were included in this research, hereafter referred to as P1-P17.  Families were recruited through regional centers within the state's Department of Developmental Services and through local school districts. Together, these two recruitment venues provide services for greater than 10,000 children and adults with ASD, with approximately 1/3 of the population under the age of ten.

Study recruitment flyers were provided to service coordinators, school district administrators, and family research center coordinators who are employed in regional centers and the schools. Families who were interested in participating in the research contacted our research team and provided  written information about their child with ASD. A licensed psychologist on our  team reviewed each child's developmental and health information for a match with the study's inclusion criteria:
\begin{enumerate}
    \item Age between 3 and 8 years old
    \item Stable physical, sensory (hearing, vision), and medical health
    \item English as a primary language spoken in the family
    \item Clinical diagnosis of ASD in mild to moderate ranges as described in the Diagnostic and Statistical Manual of Mental Disorders--Version 5 \citep{van1992comparison,kanne2008diagnostic, dover2007diagnose, baird2003diagnosis}. 
\end{enumerate}

Of the 17 children in the study, 2 were female and 15 male. They were between 3 years, 4 months and 7 years, 8 months of age. Additionally there were 3 sets of sibling pairs (P3 and P4, P5 and P6, and P16 and P17). More information about each participants living situation, education level, and age can be found in the Supplemental materials.

Due to the challenges of ASD and real-world studies, there were some exceptions among the participants. Specifically, there are no personalization data for P1 and P2, as the system was not yet fully developed for those first two deployments.  Additionally, P3 did not complete the post-study assessments for personal reasons, but did participate in the study for over a month and provided all other study data. Besides these exceptions, the rest of the participants participated in the entire study.

There is no control condition in this study, as is common in ASD studies, because individuals on the autism spectrum present an extremely broad range of symptoms, symptom combinations, and symptom severities. Consequently, work with ASD participants typically follows a single-case study model rather than the randomized trial model.  The single-case study model relies on pre/post comparisons, as was done in this paper \citep{LoboCaseStudy2017}. The pre/post WIAT Interventions in Section \ref{mm:methods:measures:objective} serve as a sample baseline to evaluate participant improvement over the course of the study. 

All child participants in the study were enrolled in full-time educational and therapeutic interventions that were consistent with the state's educational and developmental services standards and statutes. These services varied based on child needs and family preferences. All child participants had intelligence scores within ``normal'' limits levels (scores\textgreater70) based either on the Leiter International Performance Scale-3 \citep{roid2013leiter} or the by the Differential Ability Scales \citep{elliott2012differential}.

The child participants' ASD diagnoses were obtained via clinical best estimate (CBE) by trained psychologists or psychiatrists who had greater than ten years of experience in diagnosing children with ASD and other developmental disabilities. The tools used to diagnose ASD varied across clinician and referring agency. In each case, multiple measures were used to determine the diagnosis and level of ASD. Common measures used for the ASD diagnoses were the Autism Diagnostic Interview–Revised \citep{tadevosyan2003principal, wing2002diagnostic}, Autism Diagnostic Observation Schedule \citep{gotham2008replication, lord2000autism}, and Child Autism Rating Scale \citep{van1992comparison}. All children with ASD diagnoses in the study had diagnoses in the mild to moderate range \citep{van1992comparison}.

\subsection{Procedure} \label{mm:methods:procedure}
The SAR intervention was deployed in the home of each participating family for at least 30 days. The duration of each deployment was determined by when the minimum number of 20 child-robot interactions was completed; the average duration of deployment was 41 days, with a standard deviation of 5.92 days. On the day of deployment for each family, all system equipment was provided and assembled by the research team; the only requirement from participating families was a power outlet and sufficient space. During system setup, child participants were assessed by an educational psychologist using the measures described in Section \ref{mm:methods:participants}. After the system was set up, the research team conducted a system tutorial with the child participant and family.

To capture natural in-home interactions, the SAR system was fully autonomous and could be turned on and off whenever the family desired. The child participants were encouraged but not required to complete five sessions per week. Similarly, during each session, they were encouraged but not required to play each of the 10 games at least once.

\subsubsection{Objective Measures} \label{mm:methods:measures:objective}

 A large corpus of multi-modal data was collected, including video, audio, and performance on the games.   The USB camera mounted at the top of the game tablet recorded a front view of the child participant. A second camera recorded the child-robot interaction from a side view.  All interactions with the tablet were recorded, including help requests and answers to game questions.

User engagement was annotated by analyzing the camera data. A participant was considered to be engaged when paying full attention to the interaction, immediately responding to the robot’s prompts, or seeking further guidance or feedback from others in the room.

Due to numerous technological challenges common in noisy real-world studies, we were able to analyze sufficient video and audio data from seven participants (P5, P7, P9, P11, P12, P16, P17). A primary expert coder annotated whether a participant was engaged or disengaged for those seven participants. To verify the absence of bias, two additional annotators independently annotated 20\% of data for each participant; inter-rater reliability was measured using Fleiss' kappa, and a reliability of $k=0.84$ was achieved between the primary and verifying annotators.

The primary quantitative measure of cognitive skills gained throughout the study were the pre- and post- assessments, inspired by the standardized Wechsler Individual Achievement Test (WIAT II) \citep{wechsler2005wechsler} used to assess the academic achievement of children, adolescents, college students, and adults, aged 4 through 85. The test evaluates a broad range of academics skills using four basic scales: Reading, Math, Writing, and Oral Language. Within those, there are 9 subtest scores, including two math subtests, \textit{numerical operations} (NO) and \textit{math reasoning} (MR), which were the most relevant to the SAR intervention content. For young children, NO refers to early math calculations, number discrimination, and related skills; MR refers to concepts of quantity and order, early word problems, patterning, and other skills that require reasoning to solve problems. WIAT II was selected over the WIAT III because the timing of math fluency in version III presents a potential bias for children with ASD diagnoses.

The WIAT II provides raw and composite scores. Standard scores and percentile ranking are computed by comparing an individual assessment to large national samples of typically developing individuals aged 3 to adult (i.e., 2015 US normative sample $N=2,950$). A standard score of 100-110 is considered an ``average achievement score'' by national standards. The percentile ranking indicates how an individual compares to the national sample on which the tests were normed.  The WIAT-II was used as a pre-post comparison measure to determine achievement gains over the SAR intervention. Procedurally, the pre-assessment was conducted during the first few days of the intervention and the post-assessment was conducted at the end of the intervention for each child.

\subsubsection{Subjective Measures} \label{mm:methods:procedure:subjective}

We conducted biweekly interviews with participating families throughout the deployments to evaluate the system in terms of its usefulness and relationship with the participanting child, rating responses on a 7-point Likert scale with 1 being least likable and 7 being most likeable. Given the variable nature of in-home studies and different degrees of ASD across the participants, the surveys used a single-subject design \citep{horner2005use} wherein each child served as their own unique baseline. The semi-structured interviews contained similar questions, each tailored for a specific evaluation criterion, as follows.

\noindent Based on prior work by \cite{moon2001extending}, these were the questions about Kiwi's usefulness:
\begin{itemize}
    \item Does Kiwi help your child do better on the tasks? Why or why not?
    \item How could Kiwi be more useful?
    \item How involved do you have to be while your child is playing with Kiwi?
\end{itemize}

\noindent Based on prior work by \cite{rau2009effects} and \cite{Lee2005Can}, these were the questions about the child-robot relationship:
\begin{itemize}
    \item Do you think Kiwi is your friend?
    \item Do you think Kiwi listens to you?
    \item Do you feel like Kiwi knows you?
\end{itemize}
\section{Results} \label{results}
The presented month-long in-home deployments produced a large set of results. Sections \ref{results:iloc} and \ref{results:ilof} describe the patterns and quantitative results, respectively, of the hHRL framework instantiation. Section \ref{results:engagement} discusses how the adaptive system influenced the engagement of the child participants.  Section \ref{results:learning} reports on how the adaptive SAR system influenced cognitive skills gains across all participants, as measured by the pre-post intervention assessments. 

\subsection{Personalized Level of Challenge}\label{results:iloc}
\begin{figure}
    \centering
    \includegraphics[width=0.7\linewidth]{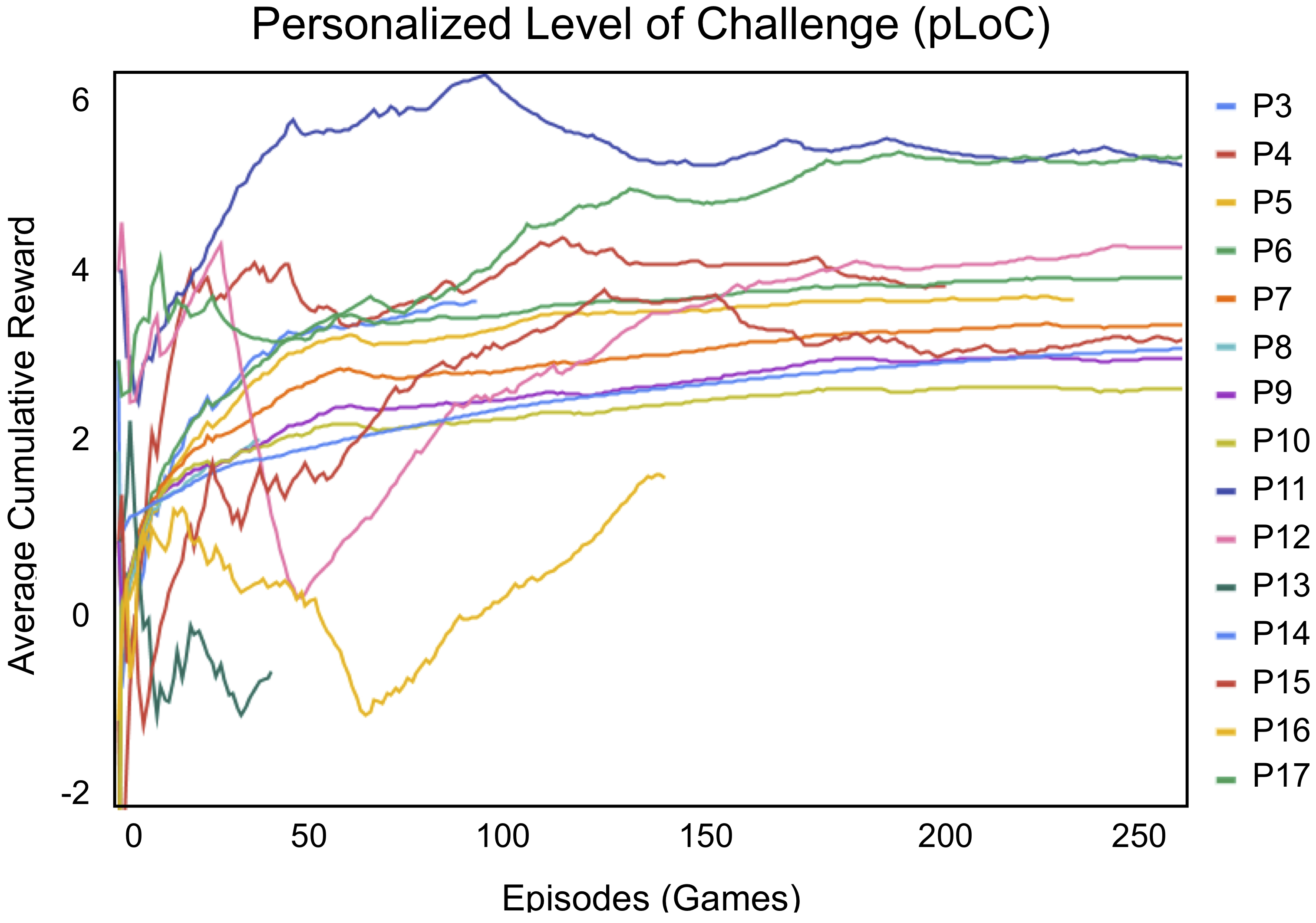}
    \caption{The reinforcement learning reward for the personalized level of challenge (pLoC) ranged between -5 and 5, where 5 indicated that the user was completing games at the highest challenge level. The average cumulative reward of pLoC matched each participant's pre-intervention scores of the WIAT II subtests for numerical operations and math reasoning. Therefore, the pLoC adapted to each participant over the month-long intervention. (There is no pLoC data for P1 or P2, as explained in Section 4.2.)} 
    \label{ilocurve}
\end{figure}

As illustrated by the learning curve in Figure \ref{ilocurve}, the personalized level of challenge (pLoC) changed over time and varied by participant. Since the goal of the adaptation was to find the optimal LoC for each participant, this learning curve cannot be interpreted in a traditional sense. For instance, if a child was not proficient at math, the learning system may not have been able to reach higher reward values, because the reward is based on both LoC and child performance.

Therefore, other factors must also be considered in interpreting the pLoC results. First, more than 100 episodes or games were required for pLoC to begin to converge. For example, for participants P3 and P8, the pLoC curve did not have a chance to converge over the few games these participants played. On the other hand, for P6, P11, and P15, the system was able to smoothly adapt given the long interaction periods. 

If a child played 10-20 games in a session, consistent with the 13.27 average of the study, then 10 sessions were required before the pLoC began to converge, totalling to approximately 132 episodes. This is a reasonable requirement given that the participants completed an average of 14.10 sessions with the robot. Excluding participants P3 and P8, who, as noted above, played significantly fewer games per session, we find an average of 17.57 sessions with the robot.

Consequently, we can conclude that the SAR system was able to adapt and personalize to each child over time.  Specifically, the pLoC implementation of the instruction controller did personalize to each child, but required a minimum number of episodes and interaction consistency to do so.

\subsection{Personalized Level of Feedback}\label{results:ilof}
\begin{figure}[t!] 
    \centering
    \includegraphics[width=0.7\linewidth]{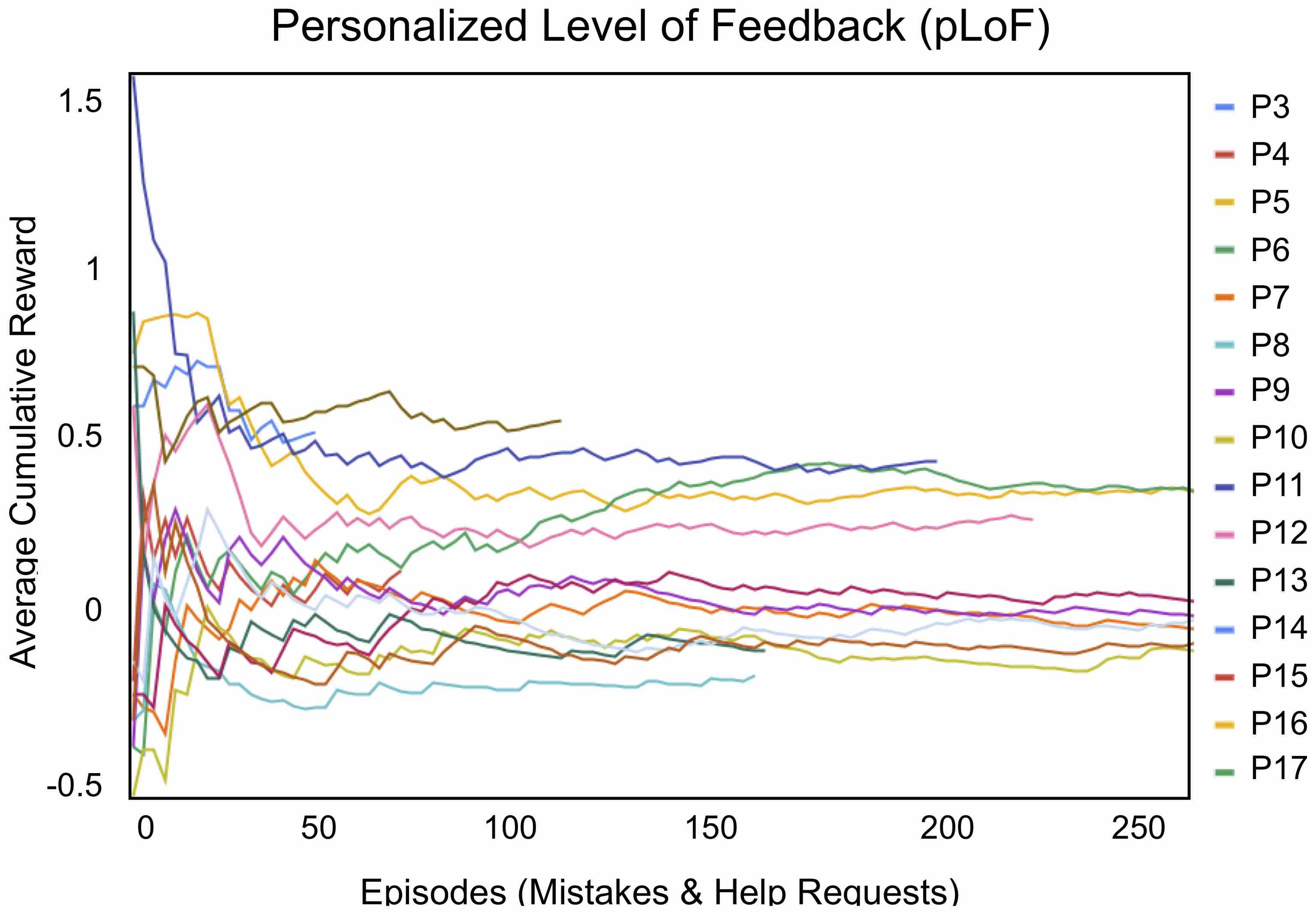}
    \caption{The reinforcement learning reward for personalized level of feedback (pLoF) ranged between 0 and 1, where 1 indicated that a child completed games with the least amount of support or feedback. The average cumulative reward of pLoF converged over 25-50 episodes (i.e., mistakes and help requests) and is correlated with the intervention length and the average number of mistakes made by child participants per game. (There is no pLoF data for P1 or P2, as explained in Section 4.2.)}
    \label{ilofcurve}
\end{figure}

The learning curve for the personalized level of feedback (pLoF) model, shown in Figure \ref{ilofcurve}, adapted the level of feedback to each participant more rapidly than pLoC. Analogous to pLoC, the pLoF learning curve cannot be interpreted in the traditional sense of simply maximizing cumulative reward; it is meant to match each child's need.

Participants with high mistake totals and long interventions usually had the longest feedback curves and, subsequently, allowed the system to adapt to their needs. The pLoF tail was longest for P10, who had the third highest mistake average, balanced with overall intervention length. Although P3 and P8 had the highest mistake averages, they also had the shortest interventions. This can be compared to the pLoF curves for P5 and P6, who had the lowest mistake averages and longest intervention lengths; the cumulative reward is higher and tails are shorter for both P5 and P6 compared to those of P10. Subsequently, P10 stands out as the longest and flattest among the three, demonstrating the value of longer interactions. Overall, the pLoF model successfully adapted to each child participant over time.

\subsection{Participant SAR Evaluation} \label{results:survey} 
SAR survey results, utilizing a seven-point Likert scale, assessed the average adaptability and usefulness of the system throughout the study.

\begin{figure}[t!]
    \centering
    \includegraphics[width=\linewidth]{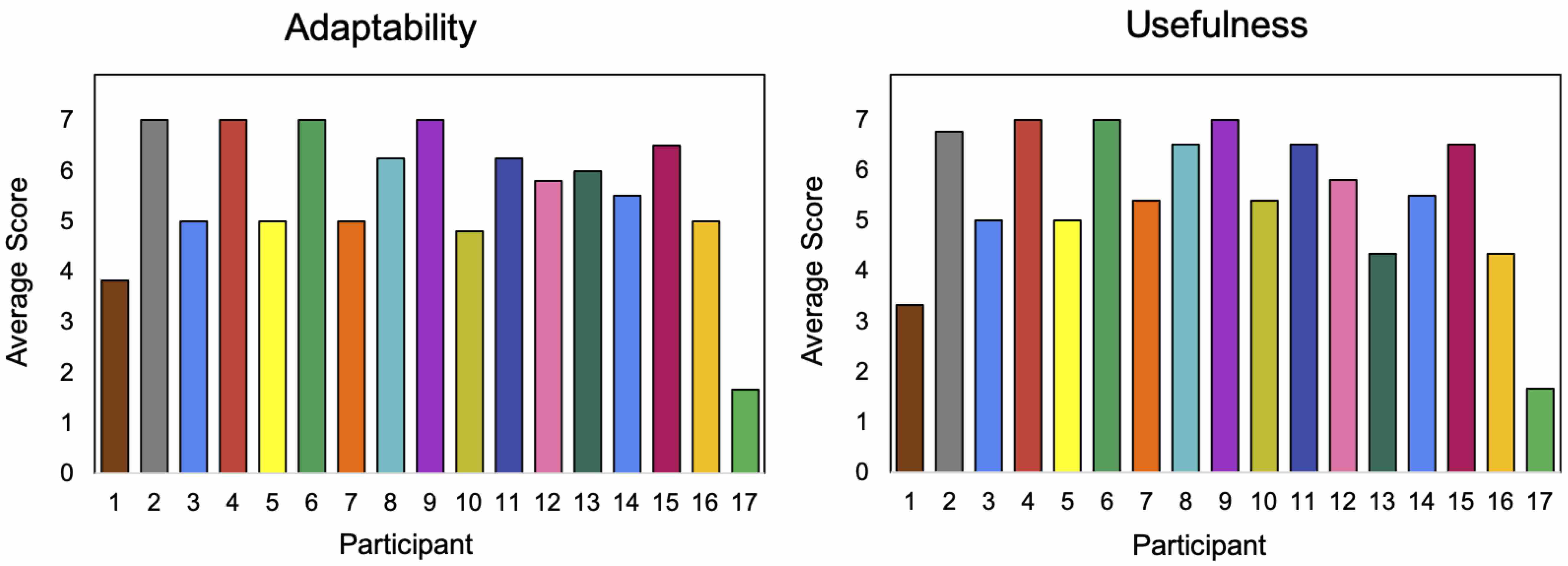}
    \caption{Participant survey results for adaptability (left) and usefulness (right).}
    \label{fig:adapt}
\end{figure}

Participants' responses about the adaptability of the SAR system, seen in Figure \ref{fig:adapt}, correlate with the challenges encountered in adapting to the individual needs of each participant. P17 reported the lowest average score for adaptability and usefulness. The result for P17 is likely due to the participant's age, as this was the second oldest and highest performing student in the study, so even the maximum difficulty was too easy for the participant. 

The reported scores for usefulness were similar to those of adaptability, as seen in Figure \ref{fig:adapt}, since the two measures are related: a system that is more adaptive to a participant is more useful. P1 and P17 were once again outliers with the lowest reported scores for usefulness. 

P1 and P7 had personal similarities: they were less than a year apart in age and had parents with the same levels of education (high school). Consequently, one would expect the system to adapt relatively similarly to both participants. Their reported scores for usefulness (5 vs 3) and adaptability (5 vs 4) were similar, thus supporting the consistency of the system across participants.

Sibling pairs (P3 and P4, P5 and P6, and P16 and P17) showed discrepancies that can be explained by the fact that the system was better suited to the needs of one sibling than the other, likely due to their age. For example, P3 was younger than P4, and therefore was not able to engage with the games as well, resulting in the lower adaptability and usefulness scores. Similarly, P6, the older sibling, reported higher scores for adaptability and usefulness than P5. The higher scores mean that the child liked the robot more and found it more adaptable and useful.

\subsection{Effect of Personalization on Engagement}\label{results:engagement}
We found that our SAR system elicited and maintained participants' engagement throughout the month-long intervention, an important measure of effectiveness. As mentioned in Section \ref{mm:methods:measures:objective}, we analyzed the seven participants (P5, P7, P9, P11, P12, P16, P17) with adequate video and audio data to analyze measures of engagement.

\subsubsection{Short-Term and Long-Term Engagement} 

\begin{figure}[t!] 
    \centering
    \includegraphics[width=\linewidth]{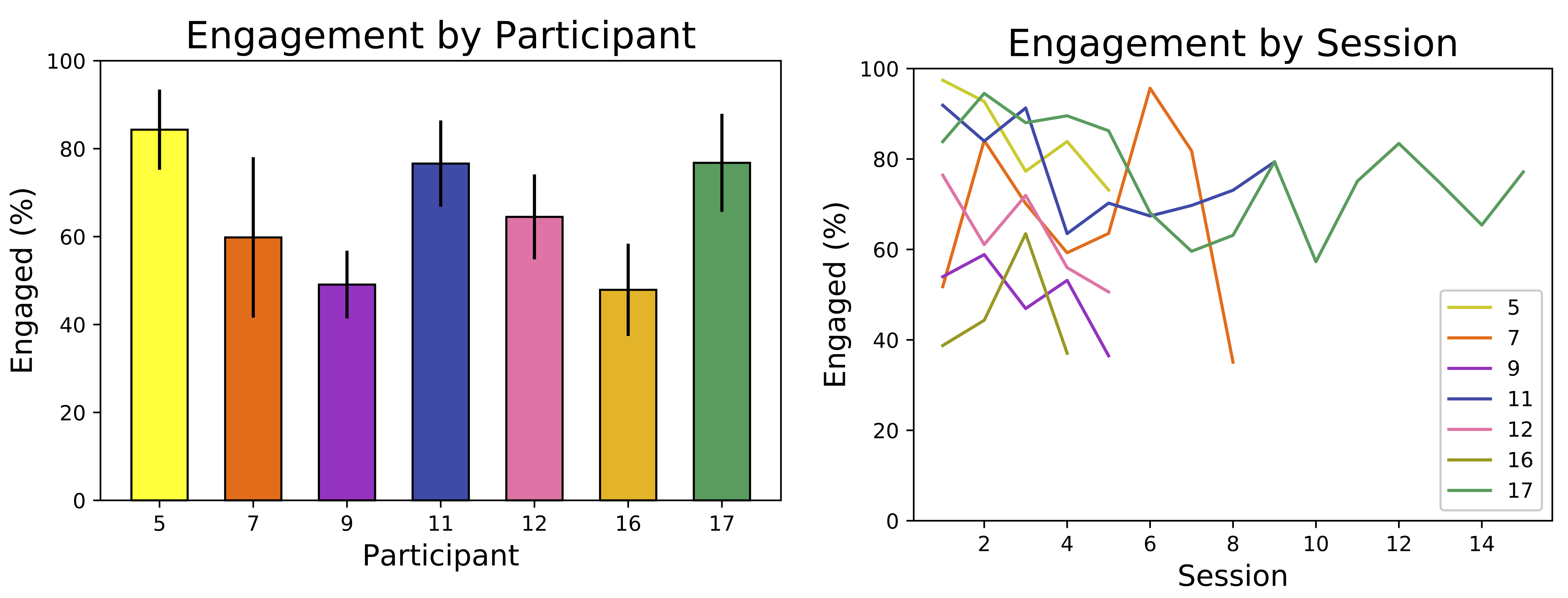}
    \caption{Overall engagement for each analyzed participant, with error bars denoting standard deviation across sessions (left). High variance but no linear trend ($p=0.99$) in engagement is observed across sessions (right). (Engagement was only analyzed for participants with adequate video and audio data.) }
    \label{eng1}
\end{figure}

The SAR system maintained reasonable levels of participant engagement during individual sessions and over the month-long intervention. As shown in Figure \ref{eng1}, all participants were engaged on average 65\% of the intervention. Across sessions, participants had an average engagement range of 32\% and standard deviation of 11\%. However, there was no statistically significant ($p=0.99$) increase or decrease in engagement over the study, as determined by a regression t-test and shown by the plotted trend line. In addition, the median duration of continuous engagement over all participants was higher than the median duration of continuous disengagement: 13 seconds to 5 seconds on average, respectively.

Furthermore, the robot was able to elicit and maintain user engagement during each game. Engagement was higher shortly after the robot had spoken; participants were engaged about 70\% of the time when the robot had spoken in the previous minute, but less than 50\% of the time when the robot had not spoken for over a minute. Participants also remained engaged after 5 minutes of starting a game nearly 60\% of the time.

\subsubsection{Engagement and the Level of Challenge}
\begin{figure}[t!] 
    \centering
    \includegraphics[width=\linewidth]{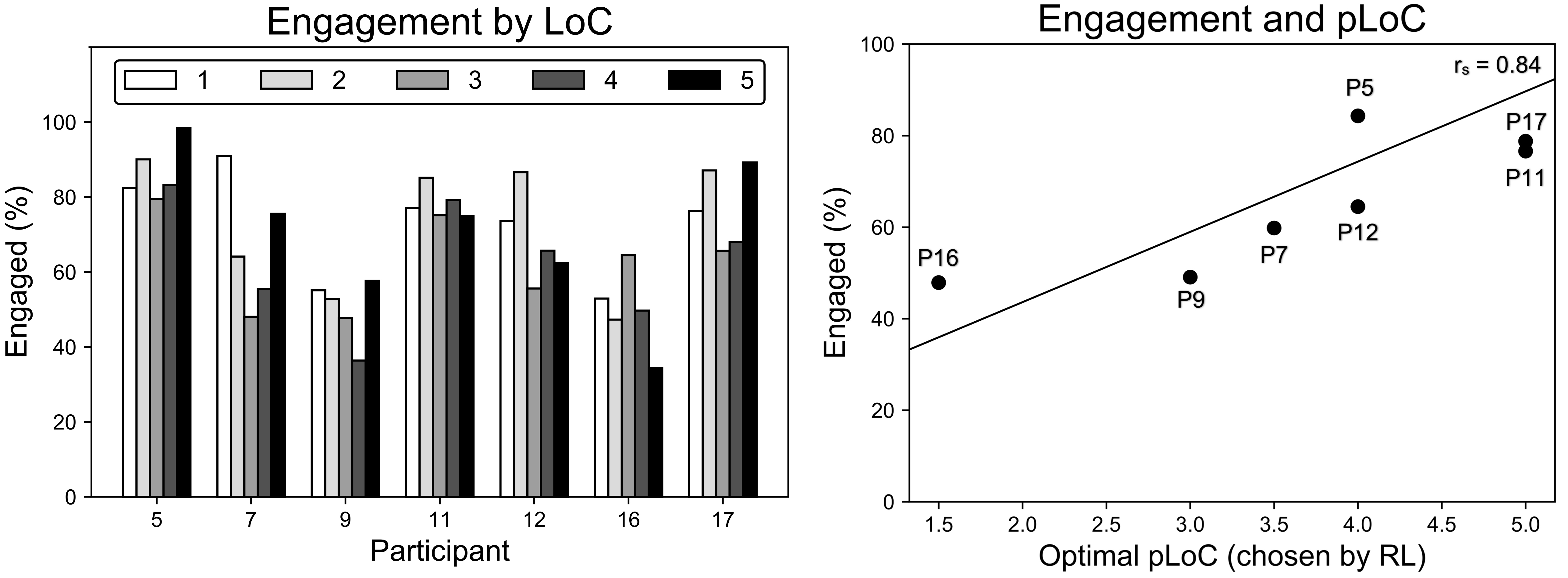}
    \caption{Variance in engagement is higher across participants than across LoC (left). Participants with high optimal LoC were more engaged ($r_{s}=0.84, p=0.018$) (right). Engagement was only analyzed for participants with adequate video and audio data.}
    \label{eng2}
\end{figure}

Engagement varied significantly across participants and their level of challenge (LoC), as shown in Figure \ref{eng2}. A two-way analysis of variance (ANOVA) showed ($p<0.01$) that average engagement for each participant varied significantly and that average engagement under each LoC also varied significantly. The variance across participants accounted for 91\% of the total variance, indicating the importance of personalization in SAR. 

The personalized level of challenge (pLoC) did not necessarily maximize engagement. As discussed above, pLoC eventually converged to an optimal LoC for each participant. But, as shown in Figure \ref{eng2}, participants whose optimal LoC was low were less engaged ($r_{s}=0.84, p=0.018$). We hypothesize that this effect is due to the time required for the learning system to adapt to each user; it took greater than 100 games for the pLoC to begin to converge, and thus participants with a lower LoC were presented with many games of higher challenge level before convergence. This further supports the importance of personalization for increasing engagement, especially with a sufficiently fast convergence rate.

\subsection{Impact on Math Learning} \label{results:learning}
\begin{figure}[t!] 
    \centering
    \includegraphics[width=\linewidth]{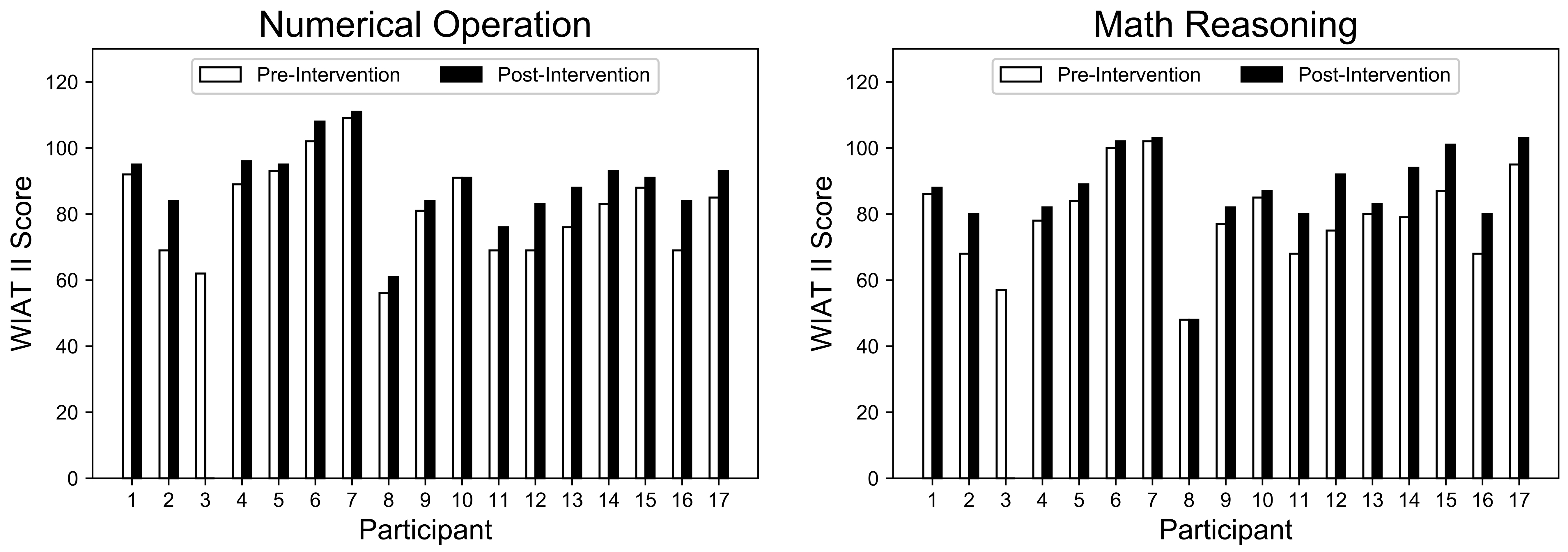}
    \caption{All participants (excluding P3 who did not complete the study) showed significant improvements ($d = 0.54$, $p<0.01$) on the WIAT II subtests for numerical operations (left), and math reasoning (right) between the pre-intervention and post-intervention assessments.}
    \label{prepost}
\end{figure}

Overall, this study observed positive gains in math learning for all participants, excluding P3 who did not complete the study. As seen in Figure \ref{prepost}, participants' pre- and post-intervention scores on the WIAT II subtests increased significantly for numerical
operations (NO) ($p = 0.002$) and math reasoning (MR) ($p < 0.001$), as determined by a t-test. In addition, both NO and MR scores had a significant effect size of $d=0.53$ and $d=0.54$, respectively, as calculated using Cohen's $d$.  

The result reveal that NO and MR both increased even when there was a large discrepancy between the initial assessment of certain participants. For example, P17 scored much higher on MR than on NO on the pre-assessment and even with such different starting points, both NO and MR increased at the post assessment. On the other hand, P11 started with the same MR and NO scores, and both scores improved after the intervention. 

When observing total cognitive gains, it is important to consider developmental factors: the age and subsequent skill level of each participant. Where older students generally had smaller net gains, they started near or above average. On the other hand, younger students started far below average, and thus had much room to improve. P8's pre-intervention scores (MR=48; NO=56) were significantly below the national average. Given P8's age (3.75 years), the scores are cautiously computed in terms of what they represent nationally. In another case,  P16's pre-intervention scores (MR=68; NO=69) were far below the national average. P16 was the youngest participant (3.11 years) and still made significant progress, improving by over 10 points in both categories (MR=80; NO=84). On the other hand, P17 was tied second oldest (7.2 years) and only made marginal gains, despite making few mistakes and performing at the highest challenge level.

\section{Discussion}
The results of the long-term in-home deployment provide several insights for personalization in SAR. 

We found that both the personalized levels of challenge (pLoC) and feedback (pLoF) converged for almost all participants. After approximately 100 games, the feedback and challenge curves stabilized, showing that the system adapted to an appropriate LoC for each student. Therefore, the long-term nature of the study was important for successful personalization. The participants with the longest episodes in the pLoC were P6, P16, and P11, with 715, 592, and 520 episodes (games played), respectively. In contrast, pLoF interacted most with for P10, P7, and P9, who had 353, 237, and 228 episodes (mistakes and help requests), respectively. The SAR system adapted to the participants in both cases. 

Regardless of the difference in sessions, participants who yielded a consistent score by the end of the interaction in the pLoC had similar success with pLoF, and vice versa. This happened for participants who interacted with equal or above average 113.4 and 302.5 episodes for pLoF and pLoC, respectively. On the other hand, P16 illustrated the negative impact of minimal interaction, as both the pLoC and pLoF failed to standardize given only 129 total interactions both on pLoC and pLoF, reaching over 171 episodes below average for pLoC. Within the interaction, the pLoC reward for P16 fluctuated by 2.6 points between the 61st and 129th episode. For reference, the second highest fluctuation in this same interval was 1.02 points by P16, whose system ultimately converged after 592 games. 

Overall, the pLoF and pLoC demonstrate the ability to adapt to each user's preferences given their willingness to interact with the robot and provide the system opportunities to learn. The participant surveys support this conclusion and provide the user's perspective on the SAR's ability to adapt.

P6, P7, P9, and P11, who both the pLoC and pLoF had adapted to, reported in their post interaction interviews an average rating of 6.25, showing a shared appreciation for the system's adaptiveness. The only study participant who believed the system did not adapt was P17, who likely felt this way because of limited success with the feedback model; P17 had only 112 feedback episodes over 298 total games. Aside from this outlier, the survey results supported the effectiveness of pLoC and pLoFs. P9 was an ideal participant, who believed the system adapted and had above average episodes while stabilizing both pLoC and pLoF. 

Usefulness questionnaire data provide additional insights into the value of creating an adaptive system. All participants reported very similar scores for usefulness and adaptiveness, implying that the usefulness of the system is related to its adaptiveness. The pre-post assessments supported this finding while providing quantitative data about the learning gains of each participant as a result of SAR personalization.

Participants whose optimal LoC was lower were less engaged, as shown in Figure \ref{eng2}. For example, the system converged to the lowest pLoC for P16, who also had the second lowest engagement. This is likely because P17 was presented with games of higher challenge before the system began to converge to an optimal LoC. When also considering that P16 had a below average number of episodes, it is likely the robot failed to adapt quickly ultimately discouraging the participant from interacting further. 


The analysis of the objective and subjective outcome measures supports the success of the system as a whole, with all participants improving in math skills over the course of the long-term in-home interaction. Regardless of whether the system was able to adapt to an optimal LoC, all participants demonstrated cognitive gains. The participants gained an average of 7.0 points on numerical operation (NO) and 7.125 points on math reasoning (MR). Although for P16 the system was unable to adapt both in pLoF and pLoC personalization, that participant was still in the top five in both NO and MO gains, with an increase of 15 and 12 points, respectively. This is likely due to the participant's initially low scores that allowed much room for improvement. All participants who had at or above average number of episodes (either in pLoC and pLoF) showed strong positive gains. P8 illustrated the disadvantages of insufficient interaction time, being the participant with the least episodes in both pLoF and pLoC and resulting with below average gains in both NO and MR assessments.

\section{Conclusion}

Socially assistive robotics (SAR) has demonstrated tremendous potential for use in high impact domains such as personalized learning for special needs populations. This work considered the problem of computational personalization in the context of long-term real-world SAR interventions. At the intersection of HRI and machine learning, computational personalization seeks to autonomously adapt robot interaction to meet the unique needs and preferences of individual users, providing a foundation for personalization. 

This work presented a formalized framework for human-robot learning as a hierarchical decision-making problem (hHRL) that decomposes a SAR intervention for tractable computational personalization, and utilized a reinforcement learning approach to personalize the level of challenge and feedback for each user. The approach was instantiated within the interactive games and tested in month-long in-home deployments with children with ASD. The SAR system wase able to personalize to the children with ASD who demonstrated cognitive gains, supporting the effectiveness of the approach.

The body of results of the presented study demonstrate that the hHRL framework and its instantiation can engage and adapt to children with diverse needs in math learning over multiple weeks. These findings highlight the tremendous potential  of in-home personalized SAR interventions.

\section*{Conflict of Interest Statement}
Matari\'c is co-founder of Embodied, Inc. but is no longer involved with the company.  Clabaugh is a full-time employee of Embodied, Inc., but was not involved with the company while the reported work was done.

\section*{Author Contributions}
Dr. Caitlyn Clabaugh developed the earning framework and its instantiation and parts of the system software,  oversaw the implementation of the rest, most of the deployments, data annotation, and provided source text for the paper.

Kartik Mahajan outlined, wrote, and edited the paper while managing various stakeholders throughout the writing process.

Shomik Jain designed and modeled the engagement process; he also detailed the Background and Engagement Results. 

Roxanna Pakkar facilitated the bi-weekly surveys and edited the Results section.

David Becerra contributed to the game performance model development, and led the first set of in-home deployments and data collections.

Zhonghao Shi designed the graphs used throughout the paper and served as a secondary editor.

Eric Deng contributed to the design of the Kiwi robot hardware and the overall SAR system for time-extended in-home deployments. 

Rhianna Lee rated the survey responses and transcribed all survey questions.

Prof. Gisele Ragusa provided domain expertise in autism and early child learning, assessment methods and tools, lead the participant recruitment, and administered the pre- and post-study assessments and interviews.

Prof. Maja Matari\'c was the project lead; she advised all students, coordinated the robot and study designs, oversaw data analysis, and extensively edited the paper.

\section*{Funding}

This research was supported by the National Science Foundation Expedition in Computing Grant NSF IIS-1139148.
\section*{Ethics Statement}
The studies involving human participants were reviewed and approved by Institutional Review Board (IRB). Written informed consent to participate in this study was provided by the participants' legal guardian/next of kin.

\section*{Acknowledgments}
The authors thank National Science Foundation Expedition Grant for funding this study along with the valuable contributions of various annotators including Balasubramanian Thiagarajan, Kun Peng, Leena Mathur, and Julianna Keller.
\section*{Supplemental Data}
 \href{http://home.frontiersin.org/about/author-guidelines#SupplementaryMaterial}{Supplementary Material} should be uploaded separately on submission, if there are Supplementary Figures, please include the caption in the same file as the figure. LaTeX Supplementary Material templates can be found in the Frontiers LaTeX folder.

\section*{Data Availability Statement}
The dataset analyzed in this study includes indentifiable video and audio data of children with autism spectrum disorders and their families, along with video information and images representing their homes. Consequently, the University IRB prohibits distribution of the dataset to protect the privacy of the research participants.

\clearpage
\bibliographystyle{frontiersinSCNS_ENG_HUMS}
\bibliography{refs}

\end{document}